\documentclass[letterpaper, 10 pt, conference]{ieeeconf}  

\IEEEoverridecommandlockouts                              

\usepackage{times}
\usepackage[pdftex]{graphicx}
\usepackage{amsmath,amssymb,amsopn,amstext,amsfonts}
\usepackage{cancel}
\usepackage[space]{cite}
\usepackage{pdfsync}
\usepackage{balance}
\usepackage{color}
\usepackage{mathtools}
\usepackage{algpseudocode}
\usepackage[ruled,vlined, linesnumbered]{algorithm2e}
\usepackage{bm}

\usepackage{diagbox}
\usepackage{float}
\usepackage{epstopdf}
\usepackage{pifont}
\usepackage{amsmath}
\usepackage{multirow}
\usepackage{url}
\usepackage{verbatim}
\usepackage[linkcolor=black,citecolor=black,urlcolor=black,colorlinks=true]{hyperref}
\usepackage{booktabs}
\usepackage{graphicx}
\usepackage{subcaption}
\usepackage{threeparttable}  
\usepackage{tikz}
\usepackage{pifont}
\usepackage{amssymb}

\makeatletter
\newcommand*{\rom}[1]{\expandafter\@slowromancap\romannumeral #1@}
\makeatother

\newcommand*{\circled}[1]{\lower.7ex\hbox{\tikz\draw (0pt, 0pt)%
		circle (.5em) node {\makebox[1em][c]{\small #1}};}}

\newtheorem{thm}{Theorem}
\newtheorem{lem}{Lemma}
\newtheorem{pro}{Problem}

\title{\LARGE \bf
Interactive-FAR:Interactive, Fast and Adaptable Routing for Navigation Among Movable Obstacles in Complex Unknown Environments
}

\author{Botao He $^{\dag \,}$\textsuperscript{1}, 
	Guofei Chen $^{\dag \,}$\textsuperscript{2}, 
	Wenshan Wang \textsuperscript{2},
        Ji Zhang* \textsuperscript{2},
        Cornelia Fermuller\textsuperscript{1},
	and Yiannis Aloimonos \textsuperscript{1}
	\thanks{\textbf{${\dag}$ Equal contribution.}}        
	\thanks{1 Perception and Robotics Group, University of Maryland, MD 20742.} 
	\thanks{2 Robotics Institute, Carnegie Mellon University, PA 15213-3890.}
	\thanks{Email: {\tt\small botao@umd.edu, \{guofei,zhangji\}@cmu.edu}}
}

\usepackage{amsmath,amssymb,graphicx}
\usepackage{xcolor}

\usepackage{tabularx}


\begin{document}

\maketitle
\thispagestyle{empty}
\pagestyle{empty}


\null\vspace{-1.0cm}

\begin{abstract}
This paper introduces a real-time algorithm for navigating complex unknown environments cluttered with movable obstacles. 
Our algorithm achieves fast, adaptable routing by actively attempting to manipulate obstacles during path planning and adjusting the global plan from sensor feedback.
The main contributions include an improved dynamic Directed Visibility Graph (DV-graph) for rapid global path searching, a real-time interaction planning method that adapts online from new sensory perceptions, and a comprehensive framework designed for interactive navigation in complex unknown or partially known environments. 
Our algorithm is capable of replanning the global path in several milliseconds. 
It can also attempt to move obstacles, update their affordances, and adapt strategies accordingly.
Extensive experiments validate that our algorithm reduces the travel time by $33\%$, achieves up to $49\%$ higher path efficiency, and runs faster than traditional methods by orders of magnitude in complex environments. 
It has been demonstrated to be the most efficient solution in terms of speed and efficiency for interactive navigation in environments of such complexity.
We also open-source our code in the docker demo\footnote{Interactive-FAR: \tt \href{https://github.com/Bottle101/Interactive-FAR}{github.com/Bottle101/Interactive-FAR}} to facilitate future research.


\end{abstract}

























\section{Introduction}

\begin{figure}
	\centering
	\includegraphics[width=0.95\linewidth]{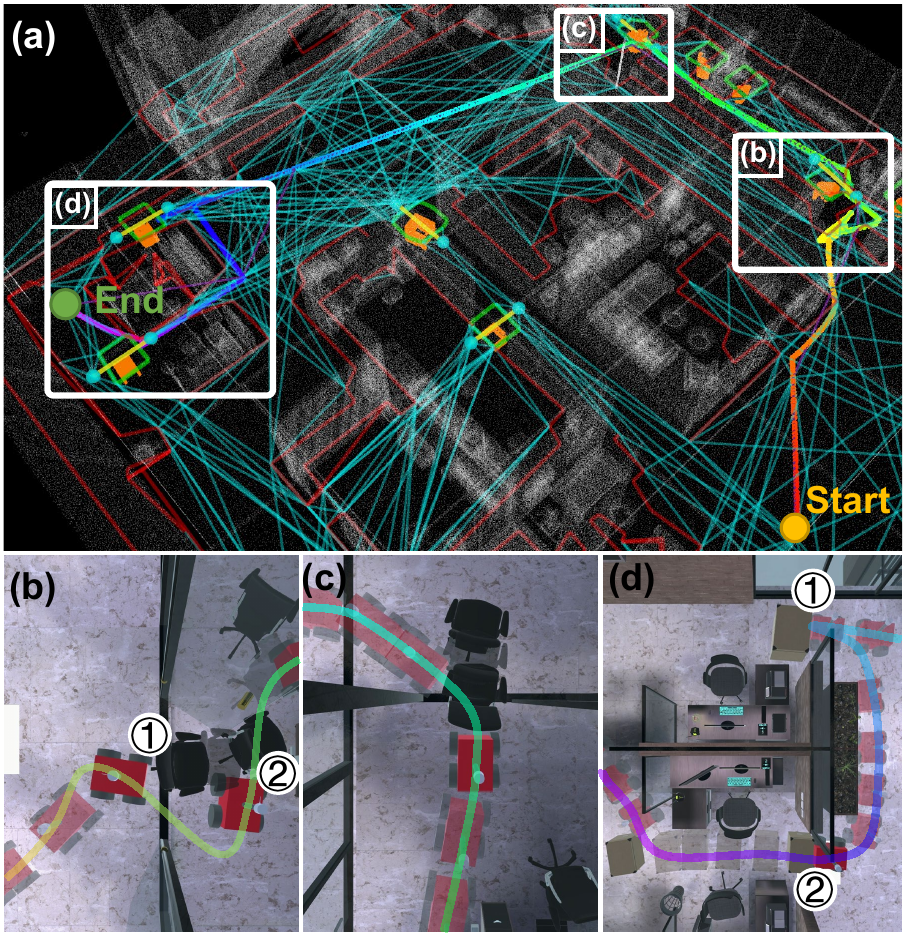}
	\captionsetup{font={small}}
	\caption{ Illustration of the interactive navigation through an unknown environment. The colorful curve is the vehicle's trajectory. The obstacles are extracted as polygons in red, and the movable objects (orange pixels) are registered as green polygons. The cyan dots represent topological waypoints. The Directed Visibility Graph is marked as cyan and yellow lines. (a) Overview. (b-c) The robot can manipulate the movable obstacles during the navigation and can switch the contact points during manipulation. 
    (d) The robot attempts to move a heavy object \textcircled{1}. When the robot contacts the object, the cost of moving it is updated from force feedback. If the object is affordable, the robot pushes it, otherwise it replans the alternate strategy to execute. The robot can quickly adapt to new sensory observations no matter whether a movable object is affordable or not.
    }
	\label{fig:cover}
	\vspace{-0.8cm}
\end{figure}

Interactive navigation is a navigation task that involves interactions between the robot and movable obstacles. Environmental interaction is essential especially for robots navigating cluttered spaces.
This paper proposes a solution for interactive navigation in cluttered unknown environments, focusing on fast and adaptable navigation with environmental interactions.
The problem remains challenging as it needs to consider multiple sub-tasks simultaneously, which include:
1) Dynamically updating the environment representation with new sensor data.
2) Executing a global path planning to avoid local minima, which considers environmental interactions during the path search. 
3) Strategizing interaction planning to manipulate objects.
In cases where the environment is complex with numerous movable obstacles, the problem's computational complexity increases significantly. Consequently, ensuring the algorithm operates in real time becomes a challenge.

Our work proposes a real-time framework that consists of identification / mapping, path planning, and interaction. This is achieved through two key ideas.

The first idea is a hybrid map representation called Directed Visibility Graph (DV-graph). 
This graph's edges represent roads and the traversability between the connected vertices. Some of the edges also encode strategies on how to manipulate movable obstacles away from the path. 
The sparse nature of DV-Graph and its integrated interaction strategies allow for a global path search with low latency (within 10ms on a 4.7GHz i7) while also considering environmental interactions. 
DV-graph is updated hierarchically at each data frame. It consists of two layers.
The local layer converts the sensor data into a polygon map, 
on which interaction planning for each movable obstacle is performed. The prospective strategies are encoded into visibility edges and connected between polygons to form the DV-graph. 
The global layer is a larger DV-graph that is merged and updated with the local layer at each frame.
The hierarchical and incremental graph construction framework ensures real-time performance during navigation.
This framework also supports manipulation actions such as push, pull, and pick-up. These can be encoded into the same map representation as a multi-level DV-graph for rapid global path search.

The second key idea is adaptable interaction planning via kinodynamic path search. 
It makes the robot actively attempt to move obstacles, update their affordances, and adapt strategies accordingly.
Given a local region divided into multiple components by a movable obstacle with unknown physics properties, as shown in Fig. \ref{fig:connectivity}, the planner tries to generate a series of kinodynamically feasible manipulation strategies that reconnect sub-regions based on assumed physical properties.
These strategies enable the robot to clear the object from its path.
During the navigation, the robot attempts to move the object by executing the above strategy, the planner updates the physics properties, thus updating the affordance, of the object through sensory feedback. The robot continues executing the original strategy if the estimated affordance is within a certain range. Otherwise, the whole global path is replanned. The updated affordance is then stored in the DV graph for later local and global planning.



The proposed algorithm is validated in complex environments of different scales and with movable obstacles. Extensive experiments show that our algorithm is faster than most benchmarks in the order of magnitudes while maintaining up to $49\%$ higher path efficiency.

The main contributions can be summarized as:
\begin{itemize}
    \item An improved dynamic DV-graph that encodes the interaction strategies and accelerates path finding.
    \item An interaction planning method that adapts to the online physics property evaluation of movable obstacles.
    \item A complete navigation algorithm for interactive navigation in unknown or partially known environments.
\end{itemize}

We open-source our code in the docker demo to facilitate future research.

\section{Related Work}

\subsubsection{Navigation in Unknown Environments}

Navigation in unknown environments has been studied and applied in many field applications, such as for exploration \cite{cao2021tare, shahidzadeh2023actexplore}, inspection \cite{farahnakian2021towards}, and swarm formation \cite{zhou2022swarm}. However, these works regard all obstacles as static and fixed, which limits the robot's ability to move in cluttered environments.


\subsubsection{Interactive Navigation - Geometry Based}
While works on navigation are prolific, there is less attention on utilizing manipulation for better navigation.
Previous studies on this problem formulate this problem as a geometric motion planning problem called navigation among movable objects (NAMO). Wilflong\cite{wilfong1988motion} proved that NAMO is NP-hard. Stilman et al.\cite{stilman2008planning} derived optimal solutions for 2D geometric planning for a subset of NAMO problems called $LP1$ with complete knowledge of the environment and objects. Levihn et al. \cite{levihn2014locally} proposed a locally optimal algorithm for $LP1$ NAMO problems in partially known environments, and Muguira-Iturral et al. \cite{muguira2023visibility} takes visibility into account in NAMO problem. 
However, solving a local NAMO problem that is moderately difficult takes a significant amount of time. In \cite{stilman2008planning}, it takes seconds due to the large search space of manipulations in a long horizon on a dense grid map. In addition, the manipulation actions are restricted to fixed contact points or directions, and the observation of object shape is assumed to be perfect. These disadvantages make them infeasible in complex environments and hinge their potential for real-world deployment.

\subsubsection{Interactive Navigation - Learning Based}
There's a recent trend for mapping visual input to agent's actions using reinforcement learning. Xia et al. \cite{xia2020interactive} compared different reinforcement learning algorithms to solve interactive navigation problems, where the robot can push the objects to shorten the path to the goal. However, the agent relies on a precomputed shortest path to operate, which is hard to achieve in real world settings. An end-to-end visual interaction navigation framework is proposed in \cite{zeng2021pushing}. However, the interaction is simplified to moving a single object  with a few predefined actions to reach a goal within its sensor range.

This work focuses on developing a real-time adaptable interactive navigation algorithm in complex environments, where all the approaches above fail. 
The introduced DV-graph based mapping and path searching algorithm facilitates rapid path planning in complex environments. 
The adaptable interaction planning algorithm empowers the robot to attempt obstacle manipulation, infer the manipulation affordance via tactile sensing, update the navigation graph, and adapt its strategies according to the latest navigation graph.



\section{Problem Definition}

\begin{figure*}[t]
	\centering
	\includegraphics[width=\textwidth]{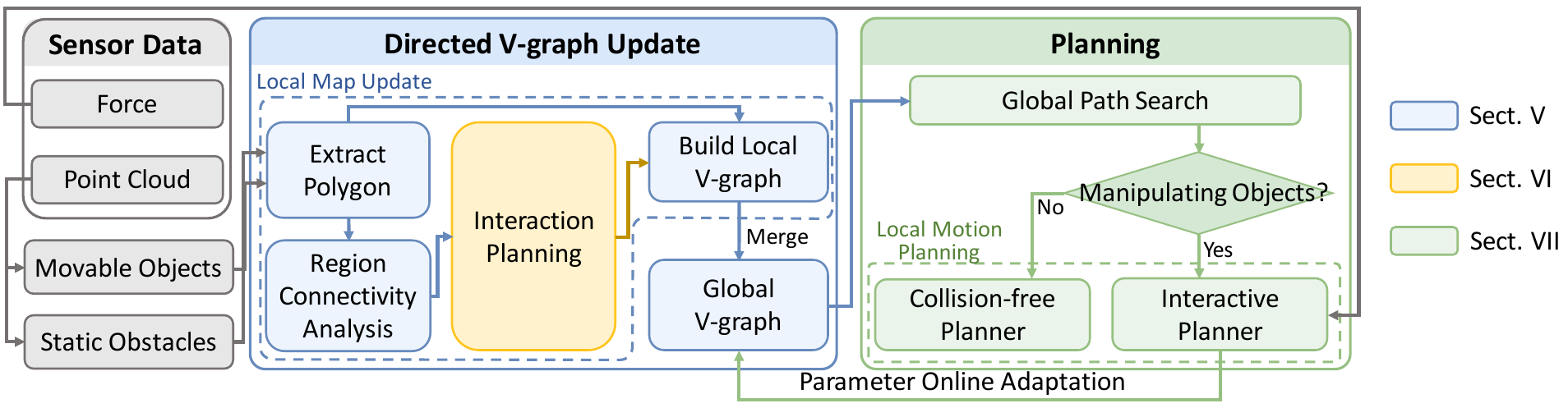}
	\captionsetup{font={small}}
	\caption{
		A diagram of our system architecture.
	}
	\label{fig:system}
	\vspace{-0.5cm}
\end{figure*}


        
Our problem is divided into two sub-problems: interactive motion planning and adaptive manipulation.

For the interactive motion planning problem, define $\textit{Q} \subset \mathbb{R}^3$ as the work space for the robot to navigate. Let $\mathcal{S} \subset \textit{Q}$ be the perceived sensor data. Here, $\mathcal{S}$ is classified as the object-level point clouds, which contains the points of movable objects $\mathcal{S}_{mov} \subset \mathcal{S}$ and points of the background obstacles $\mathcal{S}_{bg} = \mathcal{S} \backslash \mathcal{S}_{mov}$. 
We proposed a directed visibility graph (DV-graph) denoted as $\mathcal{G} = (\textit{V},\Pi)$, from $\mathcal{S}$, 
where $\mathbf{v_i} \in \textit{V}$ represents one vertex and $\langle \mathbf{v_p, v_q}\rangle  \in \Pi$ means an ordered vertex pair in $\mathcal{G}$. The problem can be expressed as:

\begin{pro}
\vspace{0.05in}
    Given the robot position $\mathbf{p}_{robot} \in \textit{Q}$ and goal point $\mathbf{p}_{goal} \in \textit{Q}$, find the most energy-efficient path between $\mathbf{p}_{robot}$ and $\mathbf{p}_{goal}$ on $\mathcal{G}$. The energy efficiency considers both the travel distance and effort efficiency when manipulating objects to move them out of the way.
\vspace{0.05in}
\end{pro}
Problem 1 is solved repetitively in each planning cycle. During navigation, we online update $\mathcal{G}$ with new sensor data, mapping the newly perceived environment and updating the maintained global map. Then, we re-plan the path on the updated $\mathcal{G}$ in the next planning cycle until arriving at $\mathbf{p}_{goal}$.

For adaptive manipulation, define $\textit{Q}_{local} \subset \textit{Q}$ as the local workspace for a single manipulation. Let $\{ \mathcal{C}^i_{local} \subset \mathcal{Q}_{local} | i \in \mathbb{Z}^+\}$ be locally disconnected components that are divided by a movable object, as shown in Fig. \ref{fig:connectivity}. The manipulation strategies are defined as motion primitives $\pi$.
The problem is described as:

\begin{pro}
\vspace{0.05in}
    Given $\textit{Q}_{local}$ that is divided into $n$ components $\{ \mathcal{C}_{local} \}$ by a movable object, find $n(n-1)$ manipulation primitives $\pi$ that re-connect any two components. 
\label{pro:problem2}
\vspace{0.05in}
\end{pro}
Problem 2 is solved once for each object that blocks a local space. The manipulation primitives are stored in the map with the object and utilized in two ways. First, it helps the DV-graph determine the travel cost from one component to another to trade off between manipulating or taking a collision-free path. Second, the searched primitives guide the robot's actual manipulation motion controller. 




\section{System Overview}


The architecture of the proposed algorithm is shown in Fig. \ref{fig:system}. The algorithm comprises two primary components: DV-graph construction and updating (Sects. V and VI), and motion planning (Sect. VII). 
In the DV-graph construction and update phase, the process begins with pre-processed sensor data $\mathcal{S}_{mov}$ and $\mathcal{S}_{bg}$. Then, a polygon extraction procedure transforms $\mathcal{S}_{mov}$ and $\mathcal{S}_{bg}$ into two sets of polygons. These sets form the map representation for the DV-graph.

The topological space connectivity analysis algorithm is designed to determine whether a single movable object divides a local space into several disconnected components.
The interaction planning module is employed for each movable object.
This module generates manipulation strategies to reconnect these locally disconnected spaces and calculates the associated manipulation costs. 
The output of the simulation module is a comprehensive set of manipulation strategies, encompassing all feasible scenarios considering both the starting and ending states.
Finally, the local DV-graph is constructed with the extracted polygons and the manipulation strategies in order to encode both the geometry and interaction information. The local DV-graph is merged with the global V-graph in each running cycle, facilitating its dynamic update.

In the planning phase, the process initially searches a global path from $\mathbf{p}_{robot}$ to $\mathbf{p}_{goal}$. 
During the path execution
, the choice of local planner depends on the robot's its current navigation context: a collision-free planner is employed when the robot navigates between two visible waypoints. An interactive planner is utilized for mobile manipulation if the current local path segment intersects with movable objects, an interactive planner is utilized for mobile manipulation. 
During the manipulation, affordance of the object are dynamically updated. The interactive planner then adaptively revises the manipulation policy based on these updated parameters. Subsequently, these revised parameters are incorporated to update the corresponding edges in the global DV-graph.


\section{DV-graph Construction and Update}

\subsection{Polygon Extraction}
\label{sec:poly_extraction}
To differentiate between static and movable objects in the DV-graph, this module takes in $\mathcal{S}_{mov}$ and $\mathcal{S}_{bg}$ separately and transform them into two sets of polygons, denoted as $\{ \mathcal{P}^i_{mov} \subset \mathcal{Q} | i \in \mathbb{Z}^+\}$ and $\{ \mathcal{P}^i_{bg} \subset \mathcal{Q} | i \in \mathbb{Z}^+\}$. To achieve that, we first inflate $\mathcal{S}_{mov}$ and $\mathcal{S}_{bg}$ based on the robot dimension, and
register them to 2-D image planes. Utilizing the methods in \cite{suzuki1985topological} and \cite{douglas1973algorithms}, enclosed polygons are extracted and simplified.
This polygon extraction step is developed from \cite{yang2022far}, detailed explanation is available therein. 
Finally, the two sets of polygons are combined into a complete local polygon graph, denoted as $\{ \mathcal{P}^i_{local} \subset \mathcal{Q} | i \in \mathbb{Z}^+\}$, as shown in Fig. \ref{fig:cover}(a). The $\mathcal{P}_{bg}$ are marked as red polygons and $\mathcal{P}_{mov}$ are marked as green.
Notably, polygons from different classes can intersect, but no intersections occur within the same class due to the topological structure analysis based on \cite{suzuki1985topological}.

\begin{figure}
	\centering
	\includegraphics[width=0.8\linewidth]{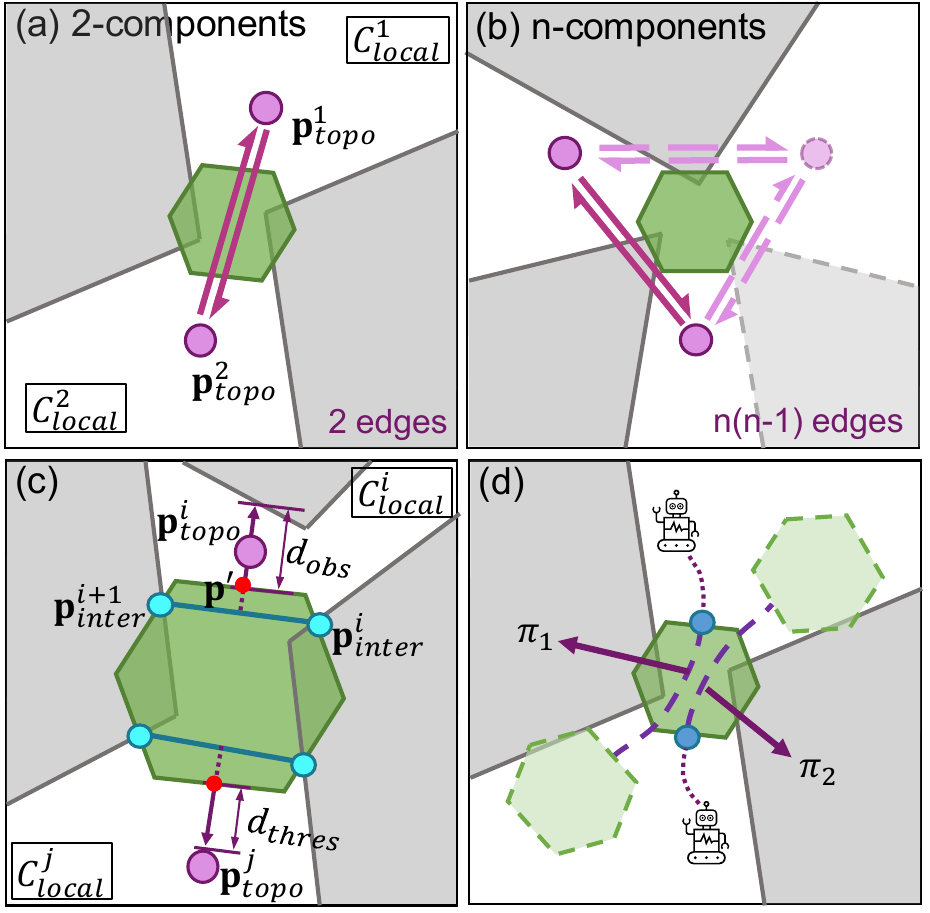}
	\captionsetup{font={small}}
	\caption{
		(a-b) Path-disconnected components $\{\mathcal{C}^i_{local}\}$ of a local region $\{ \mathcal{P}^i_{local}\}$. Green polygon represents $\mathcal{P}_{mov}$ and grey ones are $\mathcal{P}_{bg}$. They are inflated based on the robot dimension.
        Topological waypoints $\{\mathbf{p}_{topo}\}$ are marked as purple, potential manipulation strategies are represented as directed edges in purple.
        (c) Demonstration of the process to choose $\{\mathbf{p}^i_{topo}\}$. Intersected points $\mathbf{p}_{inter}$ are marked in cyan.
        (d) Illustration of the interaction planning module. Contact point are marked as blue.
	}
	\label{fig:connectivity}
	\vspace{-0.5cm}
\end{figure}

\subsection{Topological Space Connectivity Analysis}

The free-space of the static background polygon map, denote as $\mathcal{Q} \backslash \{\mathcal{P}^i_{bg}\}$, is guaranteed to be a path-connected space \cite{rodriguez2012path} because of the topological structure analysis. This implies that for any point pairs $\langle \mathbf{p}_i, \mathbf{p}_j\rangle \in \mathcal{Q} \backslash \{\mathcal{P}^i_{bg}\}$, there always exist a connected path. 
However, when $\{\mathcal{P}^i_{bg}\}$ and $\{\mathcal{P}^i_{mov}\}$ are combined, certain regions become locally path-disconnected, as shown in Fig. \ref{fig:connectivity}.
The space is divided into several locally disconnected components, $\{ \mathcal{C}^i_{local} \subset \mathcal{Q} | i \in \mathbb{Z}^+\}$.
To navigate through these disconnected spaces, the robot needs to actively manipulate the movable object to re-establish connectivity between regions.
As illustrated in Fig. \ref{fig:connectivity}(a)-(b), the number of disconnected components leads to $n(n-1)$ potential manipulation strategies based on the robot's start and goal regions.

As shown in Fig. \ref{fig:connectivity}(c), For each $\mathcal{C}^i_{local}$, define $\mathbf{p}^i_{topo} \in \mathcal{C}^i_{local}$ as the representative point of $\mathcal{C}^i_{local}$, which is called topological waypoints because different $\mathbf{p}^i_{local}$ belongs to different topological spaces.
To make topological waypoints have good visibility with other vertices in $\mathcal{C}^i_{local}$, the following strategy is adopted. 
Given a pair of consecutive intersection points $\langle \mathbf{p}_{inter}^i, \mathbf{p}_{inter}^{i+1} \rangle$, marked as cyan in Fig. \ref{fig:connectivity}(c),
calculate its mid-point $\mathbf{p}_{mid}$ and normal vector $\vec{\mathbf{n}} = \overrightarrow {\mathbf{p}_{inter}^i \mathbf{p}_{inter}^{i+1}} \times (0,0,1)$. Then create its normal vector starts from $\mathbf{p}_{mid}$, and calculate its intersection point with the polygon, denote as $\mathbf{p}'$ and mark as red point in the figure. 
Finally, 
$\mathbf{p}^i_{topo}$ can be calculated as:
\vspace{-0.2cm}
\begin{equation}
    \mathbf{p}^i_{topo} = \mathbf{p}' + \vec{\mathbf{n}} \cdot \min(d_{thres}, \frac{d_{obs}}{2}),
\end{equation}
where $d_{obs}$ represents the minimal distance from $\mathbf{p}'$ to its closest obstacle along $\vec{\mathbf{n}}$, $d_{thres}$ is a pre-defined threshold value with unit meters, here we choose $0.5$ as the $d_{thres}$.
\begin{thm}
    \vspace{0.05in}
    $\mathbf{p}^i_{topo}$ has at least 3 visible vertices in $\mathcal{C}^i_{local}$.
\end{thm}

\textit{Proof: }To prove it, two axioms are introduced: i) Any polygon can be decomposed into triangles. ii) Any point in a triangle is visible to all three vertices. Given $\mathcal{C}^i_{local}$, it can be decomposed to a set of triangles $\{\Delta_k \subset \mathcal{C}^i_{local} | k \in \mathbb{Z}^+\}$, because $\mathbf{p}^i_{topo}$ is within $\mathcal{C}^i_{local}$, it must be in one of $\{\Delta_k\}$, thus connecting at least three vertices in $\mathcal{C}^i_{local}$. 

\vspace{0.2cm}

If the robot can manipulate the object from $\mathcal{C}^i_{local}$, and achieves $\mathcal{C}^{j}_{local}$, then the point pair $\langle \mathbf{p}^i_{topo}, \mathbf{p}^{j}_{topo} \rangle$ is connected added into the DV-graph as a travelsable edge. The travelsibility of the edge is assigned based on the estimated cost of the manipulation.

\setlength{\textfloatsep}{0pt}
\begin{algorithm}[t!]
\SetAlgoLined
\LinesNumbered
\SetKwInOut{Input}{Input}
\SetKwInOut{Output}{Output}
\Input{Segmented Sensor Data: $\mathcal{S}_{bg}, \mathcal{S}_{mov}$, DV-graph: $\mathcal{G}$}
\Output{Updated DV-graph: $\mathcal{G}$}
$\{ \mathcal{P}_{bg}, \mathcal{P}_{mov} \} \leftarrow $ PolygonExtraction($\mathcal{S}_{bg}, \mathcal{S}_{mov}$)\;

\Comment{Local DV-graph update}

Construct local DV-graph on $\{ \mathcal{P}^i_{bg} \}$\;
\For{each $\{ \mathcal{P}_{mov} \}$}{
\For{each pair $\langle \mathbf{p}_{topo}^i, \mathbf{p}_{topo}^{j} \rangle$}{
$\mathcal{J},\pi \leftarrow$ $\Gamma$($\langle \mathbf{p}_{topo}^i, \mathbf{p}_{topo}^{j} \rangle$, $\mu$)
}
}

\Comment{Local-global DV-graph merge}

\For{each vertex $ \mathbf{v}_p \in \{ \mathcal{P}^i_{local} \} \cup \{ \mathcal{P}^j_{global} \}$}{
\uIf{an association exists}{
Update the vertex in $\{ \mathcal{P}^j_{global} \}$\;
}
\uElseIf{$\mathbf{v}_p$ only exists in $\{ \mathcal{P}^j_{local} \}$ }{
Add $\mathbf{v}_p$ to $\{ \mathcal{P}^j_{global} \}$ as a new vertex\;
}
\Else{
Remove $\mathbf{v}_p$ from $\{ \mathcal{P}^j_{global} \}$ based on voting\;
}
}

\Return Merged DV-graph $\mathcal{G}$\;

\caption{DV-graph Update}\label{alg:updategraph}
\end{algorithm}

\subsection{Interaction Planning}
\label{sec:simulation}
Given a locally path-disconnected region with n-components, as shown in Fig. \ref{fig:connectivity}(a)-(b), the module is designed to generate $n(n-1)$ local manipulation policies that make each $\langle \mathbf{p}^i_{topo}, \mathbf{p}^{j}_{topo} \rangle$ re-connected.
The input of this module is the local contour graph $\{ \mathcal{P}_{local} \}$. For each movable object $\{ \mathcal{P}^i_{mov} \subset \mathcal{P}_{local} \}$, the simulation process is described in Algorithm \ref{alg:updategraph}, Line 5-9. Given an arbitrary pair of topological waypoints $\langle \mathbf{p}^i_{topo}, \mathbf{p}^{j}_{topo} \rangle$ with the estimated push affordance $\mu$ of the object, the strategy generation function $\Gamma: \mathbb{R}^2 \times \mathbb{R}^2 \rightarrow \mathbb{R} \times \mathbf{SE(2)}$ outputs
the interaction strategies $\pi$ along with the estimated manipulation cost $\mathcal{J}$. 
Because the output does not restrict manipulation actions, general real-time mobile manipulation solutions can be conveniently adopted into this framework with minor changes.
The detailed design of the interaction strategies generation is discussed in Section \ref{sec:prigen}.

\subsection{Local-Global DV-graph Update}
After constructing $\{ \mathcal{P}_{local} \}$, we merge the overlapped area of $\{ \mathcal{P}_{local} \}$ and $\{ \mathcal{P}_{global} \}$ to update the global DV-graph $\mathcal{G}$. 
The process is described in Algorithm \ref{alg:updategraph}, Line 11-19. For each vertex in $\{ \mathcal{P}_{local} \}$, we check if there's a close match in $\{ \mathcal{P}_{global} \}$ within a certain distance threshold. If so, we update the existing vertex in $\{ \mathcal{P}_{global} \}$. 
If the vertex only exists in $\{ \mathcal{P}_{local} \}$, we regard it as a new vertex and add it into the $\{ \mathcal{P}_{global} \}$.
Conversely, vertices only in $\{ \mathcal{P}_{global} \}$, which means they are not observed in the current frame, are not immediately removed. 
To mitigate sensor noise, we develop a voting mechanism where a vertex is removed from $\{ \mathcal{P}_{global} \}$ only after being continuously unobserved for several frames.

\section{Interaction Strategy Generation}
\label{sec:intergen}
In this section, we propose our solution for Problem \ref{pro:problem2}. We first decompose the Problem \ref{pro:problem2} into a series of sub-problems (Problem \ref{pro:subproblem}). For each sub-problem, we conduct a kinodynamic path search for a feasible interaction strategy by expanding motion primitives with a discretized control input space (Sect. \ref{sec:prigen}). Our novel design of heuristics is proven to be admissible in general cases and fits well with our sparse polygon-based map representation.

\subsection{Problem Decomposition}
As discussed in Sect. \ref{sec:simulation}, there are $n(n-1)$ local policies to be generated. It is hard to solve via a single search because the state space is large and with multiple sub-goals, and it is hard to design heuristics with an admissibility guarantee. Therefore, we decompose the original problem into $n(n-1)$ sub-problems and solve them separately. 

\begin{figure}
	\centering
	\includegraphics[width=0.95\linewidth]{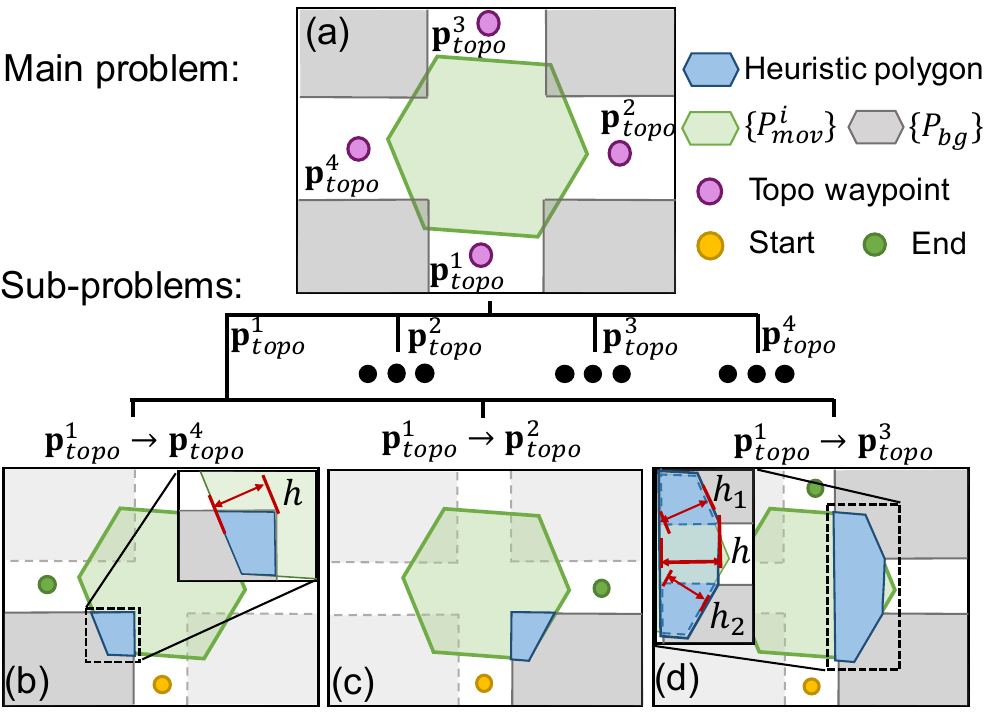}
	\captionsetup{font={small}}
	\caption{
		Illustration of the problem decomposition. (a) The main problem contains $n(n-1)$ sub-problems. (b-d) Three typical sub-problems. The upper-right corner of (b) explains the calculation of the heuristics. The left-upper corner of (d) demonstrates the case when there are multiple heuristic polygons.
	}
	\label{fig:divide}
	\vspace{-0.3cm}
\end{figure}

The process of the problem decomposition is illustrated in Fig. \ref{fig:divide}. Each sub-problem is defined as:
\begin{pro}
\vspace{0.05in}
    Given the start position $\mathbf{p}^i_{topo}$ and goal position $\mathbf{p}^j_{topo}$, find the most energy-efficient manipulation policy that connects $\mathbf{p}^i_{topo}$ and $\mathbf{p}^j_{topo}$ via a local collision-free path.
\label{pro:subproblem}
\vspace{-0.08in}
\end{pro}

The problem is solved by a hybrid A* path search method with a novel heuristic design. It has two advantages: 1) the found interaction strategy, or path, is kinodynamically feasible for the robot to execute, and 2) the designed heuristics are admissible in general cases, which improved the optimality of the path. 
Technical details can be found in Sect. \ref{sec:hybridastar}.


\subsection{State Transition Model}
\label{sec:prigen}
In this section, we discuss the state transition modeling in the kinodynamic path search algorithm. For simplicity of our system, we use differential vehicles: $v = [v_x, 0], \omega \in \mathbb{R}$, and use push manipulation as our interacting method.

Due to the sensor limitation of mobile robots,
our system only infer the object's shape from on-board depth input (where occlusion is inevitable) and physics properties by push attempts. For this reason, we choose stable push\cite{lynch1996stable} as our scheme. It's more robust to the estimation error of object's shape and physics properties compared with other push schemes\cite{mason1986mechanics, hogan2018reactive, doshi2020hybrid}. It also simplifies the generation of kinodynamically feasible push strategies for our vehicle.


Given an object with known shape, friction center, contact edge, and the robot to object friction coefficient, we can identify a set of potential stable rotation centers $P_s$  by the procedure "STABLE" in \cite{lynch1996stable}. In our case, the object shapes and physics properties are unknown at first. We estimate the shape of the object in \ref{sec:poly_extraction}, and start with an initial robot to object friction coefficient \(k\). It is assumed to be uniform so that the friction center aligns with its geometric center.
Then, for a certain contact edge (which determines the robot state), we select the vehicle's control inputs which will only result in stable pushes of the object, and map them to the state of objects. Such mapping is trivial because of stable push. 

For our case, the full state space $\mathbb{S'}$ of problem 2 is the combination of the robot's and object's states: $\mathbb{S'} \subset \mathbf{SE(2)} \times \mathbf{SE(2)}$. 
Because of the bijection between the robot state and the object's contact points, the robot's state can be simplified as a set of available contact points $\mathbf{C} \subset \mathbb{R}^2$, shown as blue points in Fig. \ref{fig:connectivity}(d). The state space is then simplified to $\mathbb{S}\subset \mathbf{C} \times \mathbf{SE(2)}$.
The state vector of a movable object is comprised of object position and orientation $x = [p_x, p_y, \psi]^T$ and available contact points $c = [c^1, c^2, ..., c^n]^T$, where $c^i \in \mathbf{C}$. The state transition model is then derived by: a) applying the differential vehicle model to the movable object and aligning the heading with the normal of the stable push edge. b) restrictig $\frac{\omega}{v_x}$ to let the instantaneous rotation center $p$ fall into $P_{s}$. A state transition model whose actions are feasible and transitions are robust is generated.
The physics properties will update during the execution of push strategies, which we discuss in VII.B.

\begin{figure}[t!]
	\centering
	\includegraphics[width=0.95\linewidth]{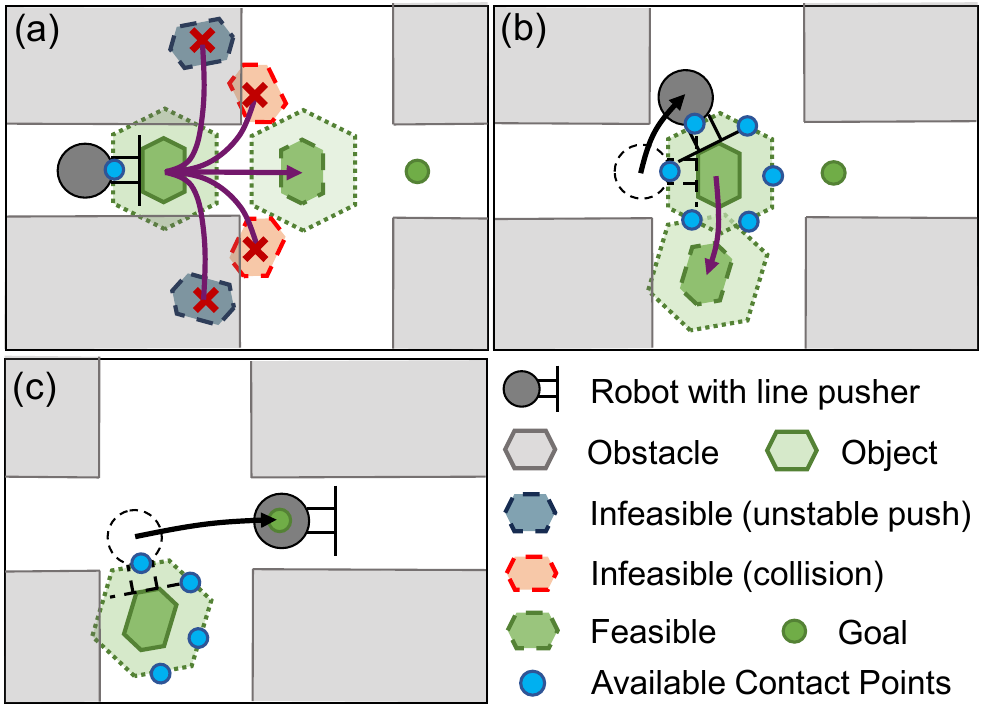}
	\captionsetup{font={small}}
	\caption{
		Hybrid A* push search with contact point switch. 
	}
	\label{fig:astar}
	\vspace{-0.4cm}
\end{figure}

\subsection{Hybrid A-Star Path Search}
\label{sec:hybridastar}
Given a pair of topological waypoints, the problem \ref{pro:subproblem} is solved in two steps. 
First, find the polygon with the smallest area that hinders the connection of these topological waypoints, demonstrated as blue polygons in Fig. \ref{fig:divide}(b-d). Here we call it heuristic polygon because it provides heuristics to guide the path search. We represent it as $\{\mathcal{P}_{heu}\} = \{\mathcal{P}^i_{mov}\} \cap \{\mathcal{P}_{bg}\}$.
If there are multiple heuristic polygons between the two regions, $\{ \mathcal{P}_{heu}\}$ is chosen as the union of the vertex set of all these polygons, as shown in Fig. \ref{fig:divide}(d).
The heuristic (or the negative reward) of the path search is the minimum distance between parallel lines of the polygon, which is mathematically defined to be the width of the polygon\cite{houle1985computing}. 
A hybrid-state astar is then performed with the state and transition model discussed in Sect. \ref{sec:prigen}.  


\subsection{Completeness and Admissibility}


The completeness cannot be theoretically guaranteed because the state space is discredited. However, our method supports variable-resolution state space to improve the completeness guarantee \cite{dolgov2010path} and the practical performance is satisfactory.
For optimality, we can prove the admissibility when the heuristic polygon $\{\mathcal{P}^i_{heu}\}$ or the union of all vertices of multiple heuristic polygons $\{\mathcal{P}^{union}_{heu}\}$ are convex.

\begin{lem}
\vspace{0.05in}
  $h$ is admissible if $\{\mathcal{P}^i_{heu}\}$ is convex.
\vspace{0.05in}
\end{lem}
According to the definition stated in \cite{houle1985computing}, $h$ is the minimum distance that disengage two convex polygons from contact. Therefore, $h$ is admissible.

\begin{thm}
\vspace{0.05in}
  $h$ is admissible if $\{\mathcal{P}^{union}_{heu}\}$ is convex.
\vspace{0.05in}
\end{thm}

{\it~~Proof}: As shown in the upper-left part of Fig. \ref{fig:divide}(d), if the union is convex, $h$ must satisfy the condition that $h \leq h_1 + h_2$. Because $h_1$ and $h_2$ are both admissible for their heuristic polygons, therefore $h$ is admissible for the union.

\section{Interactive Motion Planning and Online Adaptation}

\subsection{Global Path Search on DV-graph}

To identify the DV-graph $\mathcal{G}$ for the shortest path from $\mathbf{p}_{robot}$ to $\mathbf{p}_{goal}$, the planner first add $\mathbf{p}_{robot}$ and $\mathbf{p}_{goal}$ as two vertices, connecting them to other visible vertices in ${P^i_{global}}$. A breadth-first search is then executed on $\mathcal{G}$ to find the shortest path, denoted as a set of waypoints $ \mathbf{P}_{path} = \{ \mathbf{p}^i_{wp} \in \mathcal{Q} | i \in \mathbb{Z}^+\}$, if it exists.
During the navigation among the unknown environment, vertices and directed visibility edges are dynamically updated and stored in $\mathcal{G}$. In the subsequent operations, the DV-graph $\mathcal{G}$ can be loaded as a prior map. When the environment is fully explored, the global planner is capable of efficiently conducting real-time, globally optimal path searches in complex environments ($455m$ path with $1570$ vertices in $3ms$, as tested).


\subsection{Path Execution and Adaptable Replan}

During the navigation, the robot switches the local planner based on the current path segment it is executing. The working principle is demonstrated in Algorithm \ref{alg:localplanner}.
The robot employs collision-free planner \cite{cao2022icra} for regular navigation.
Upon nearing a topological waypoint, it switches to interactive planner to execute the planned interaction strategies discussed in Sect. \ref{sec:intergen}. 
The interactive planner considers the following constraints during the runtime: push stability, collision avoidance and push affordance $\mu$ of the robot (e.g., how much weight can it push, the friction coefficient between pusher and slider). We utilize the same method discussed in Sect. \ref{sec:prigen} for path replan.

With each sensor frame, the robot utilizes tactile sensing to update the physical properties of the movable obstacle and therefore update the push affordance.
The design of our tactile feedback module is based on three principles:
\begin{enumerate}
    \item If the state of the object doesn't change after maximum push effort, we mark the object as not pushable.
    \item If $||\mathcal{J} - \hat{\mathcal{J}}||$ exceeds a threshold, update $\mathcal{J}$ and $u_x$.
\end{enumerate}

Based on the updated $\mathcal{J}$ and $u_x$, the planner adaptively re-plans the interaction primitive or the global path as necessary. After the push execution, the updated $\mathcal{J}$ and $\pi$ are encoded into the DV-graph $\mathcal{G}$.
No matter whether the object is affordable, the proposed method can always adapt to the affordance update and replan the proper strategy accordingly.

\setlength{\textfloatsep}{-15pt}
\begin{algorithm}[t!]
\SetAlgoLined
\LinesNumbered
\SetKwInOut{Input}{Input}
\SetKwInOut{Output}{Output}
\Input{
Next waypoint: $\mathbf{p}_{path}^{next} \in \mathbf{P}_{path}$, \\
start: $\mathbf{p}_{robot}$, goal: $\mathbf{p}_{goal}$, cost: $\mathcal{J}$, \\affordance: $\mu$
}

\While {$\mathbf{p}_{robot} \neq \mathbf{p}_{goal}$} {
\uIf{$\mathbf{p}_{wp}^{next} \in {\mathbf{p}_{topo}}$ and 
$||\mathbf{p}_{wp}^{next} - \mathbf{p}_{robot}|| < r$
}{
$\hat{\mathcal{J}}, \hat{\pi}, \hat{\mu} \leftarrow \Gamma (\langle \mathbf{p}_{robot}, \mathbf{p}_{wp}^{next} \rangle, \mu)$;

\uIf{
$||\hat{\mathcal{J}} - \mathcal{J}|| > \tau_{thres}$
}{
$\mathcal{J} \leftarrow \hat{\mathcal{J}}$; $\pi \leftarrow \hat{\pi}$; $\mu \leftarrow \hat{\mu}$; \\
Update correlated edges in $\mathcal{G}$ based on $\hat{\mathcal{J}}$; \\
Replan the global path;
}
\Else{

InteractivePlanner($\pi$);

}

}
\uElse{
CollisionFreePlanner($\mathbf{p}_{wp}^{next}$);
}
}

\caption{Path Execution and Adaptation}\label{alg:localplanner}
\end{algorithm}

\section{Experiments}

\begin{figure}[t!]
	\centering
	\includegraphics[width=0.95\linewidth]{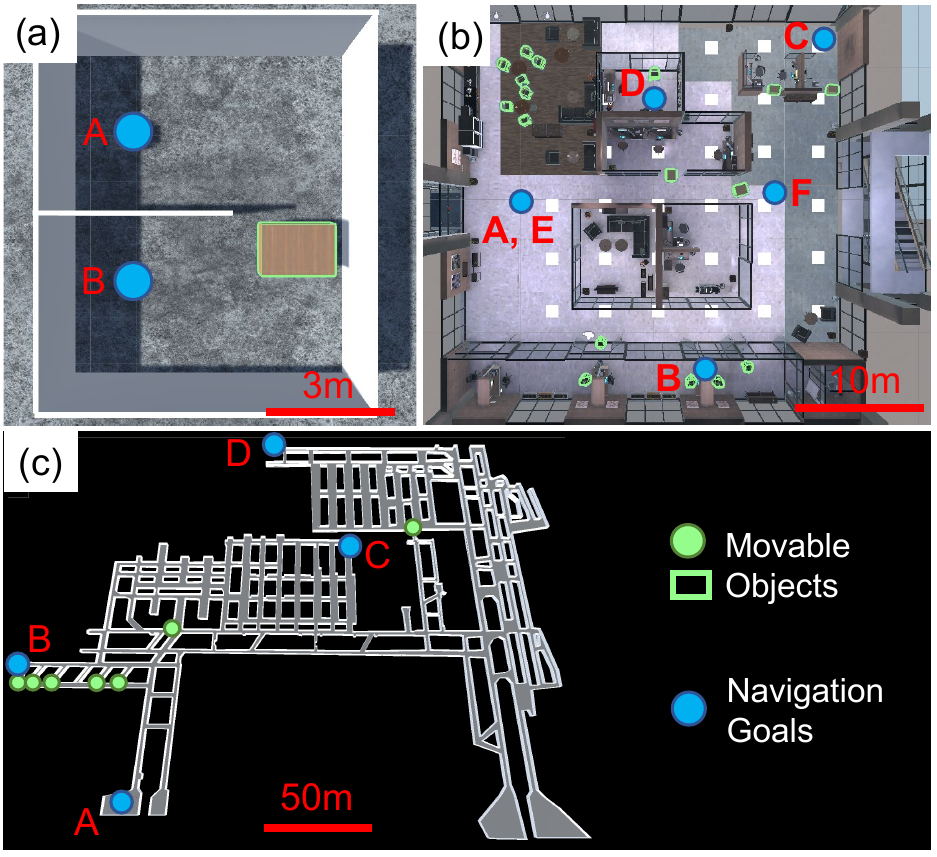}
	\captionsetup{font={small}}
	\caption{
		Environments overview. 
	}
	\label{fig:env}
	\vspace{0.5cm}
\end{figure}

To demonstrate the effectiveness of our interactive navigation algorithm, we conduct $10$ navigation tasks in simulated environments with different scales, as shown in Fig. \ref{fig:env}. 
For each task, The start and goal position are chosen based on the criteria that the optimal path length is larger than a threshold. The threshold is set as $15m$ for the $32m \times 32m$ environment, and $80m$ for the $330m \times 270m$ one.
The simulated ground vehicle is equipped with a Velodyne Puck Lidar used as the range sensor for perception, a 1-D force sensor for tactile sensing. 
The framework runs on a laptop with i7-12700H CPU. We configure the algorithm to update the DV-graph at 2.5Hz and perform path search at each graph update. The spatial resolution is set as $0.15m$. The local layer is a $60 \times 60 m$ area with the vehicle in the center.

\subsection{Computational Efficiency Comparison}
\label{sec:comptime}

To demonstrate the computational efficiency of the proposed framework, we compare our algorithm with three classical path planning methods: \textbf{A*}, \textbf{RRT*} and \textbf{BIT*}, two NAMO methods: \textbf{R-NAMO} and \textbf{LAMB}, and our previous work \textbf{FAR} that uses a DV-graph without considering interaction. 
Here, A* is a representative search-based method, BIT* is considered as the state-of-the-art in sampling-based method, R-NAMO and LAMB are two representative methods for NAMO in known and unknown environments, and FAR is considered as the latest global path planning method in unknown environments.
The experiment results are demonstrated in Table. \ref{tab:time}.
The italic font indicates that the algorithm is implemented in Python, otherwise implemented in C++.
The ground truth (GT) optimal path is noted as the length of the path search by A* on the dense grid map without inflation.
The average time consumption is recorded in Table. \ref{tab:time} if the path search is successful.
The time recorded for sampling-based methods represents the duration until the method initially finds a sub-optimal path, the length of which is no more than 1.5 times the length of the GT optimal path.

As Table. \ref{tab:time} shows, considering that the speed of Python can be up to $96.3$ times slower than C++ \cite{zehra2020comparative}, all methods have impressive performance in the room environment. 
In the office scenario, search-based method runs slower due to the increase of the environment scale, while all NAMO methods fail in this scale. 
In the tunnel scene, Only FAR and our method can run in real-time and maintain similar speeds.
The reasons for the performance of our work and FAR can be attributed to 1) the sparse nature of the map representation makes the path search faster in orders of magnitude, and 2) the efficiency of C++ makes the algorithm runs faster.

\subsection{Path Efficiency Comparison}

To demonstrate the path efficiency of the proposed system, we conducted this experiment using the same experiment settings as Sect. \ref{sec:comptime}. 
Table. \ref{tab:length} presents the results of the experiments. The performance of path efficiency is evaluated by SPL\cite{anderson2018evaluation}. As shown in the table, A* performs best in the room scenario because of its mapping density and optimality guarantee.
In the room with movable obstacles, R-NAMO and LAMB have lower SPL because their manipulation policies take redundant actions. However, our method can plan high quality manipulation policies because it is admissible in general cases. 
In Office and Tunnel, with the increasing scale of the environments, the path length of manipulation actions has a lower ratio than the total path length; therefore, our method achieves higher SPL in these scenarios. 
However, R-NAMO and LAMB fail due to their low scalability, and other methods have to take alternate paths or even fail in some sub-tasks as they cannot conduct interactive navigation, thus resulting in lower SPL.

\begin{table}[t!]
\centering
\caption{Average Search Time in [ms]}
\label{tab:time}
\begin{tabular}{|ccccc|}
\hline
Environment                                                       & \begin{tabular}[c]{@{}c@{}}Room\\ (no objects)\end{tabular} & \begin{tabular}[c]{@{}c@{}}Room\\ (with objects)\end{tabular} & Office                            & Tunnel        \\ \hline
\begin{tabular}[c]{@{}c@{}}Optimal\\ Path Length (m)\end{tabular} & 8.6                                                         & 8.6                                                           & 47                                & 455           \\ \hline
\multicolumn{1}{|c|}{A*}                                          & \multicolumn{1}{c|}{2.2/\textit{179.0(py)}}                              & \multicolumn{1}{c|}{-}                                     & \multicolumn{1}{c|}{16.0}         & 363.5         \\ \hline
\multicolumn{1}{|c|}{RRT*}                                        & \multicolumn{1}{c|}{0.4}                                    & \multicolumn{1}{c|}{-}                                      & \multicolumn{1}{c|}{\textbf{0.4}}          & 325.2         \\ \hline
\multicolumn{1}{|c|}{BIT*}                                        & \multicolumn{1}{c|}{\textbf{0.17}}                          & \multicolumn{1}{c|}{-}                            & \multicolumn{1}{c|}{0.7}          & 1.1e3         \\ \hline
\multicolumn{1}{|c|}{R-NAMO}                                        & \multicolumn{1}{c|}{\textit{1.1e3(py)}}                         & \multicolumn{1}{c|}{\textit{1.2e3(py)}}                           & \multicolumn{1}{c|}{-}            & -             \\ \hline
\multicolumn{1}{|c|}{LAMB}                                        & \multicolumn{1}{c|}{\textit{7.0e3(py)}}                         & \multicolumn{1}{c|}{\textit{34.8e3(py)}}                          & \multicolumn{1}{c|}{-}            & -             \\ \hline
\multicolumn{1}{|c|}{FAR}                                         & \multicolumn{1}{c|}{0.3}                                    & \multicolumn{1}{c|}{-}                                      & \multicolumn{1}{c|}{\textbf{0.4}} & \textbf{1.27} \\ \hline
\multicolumn{1}{|c|}{Ours}                                        & \multicolumn{1}{c|}{0.3}                                    & \multicolumn{1}{c|}{\textbf{0.3}}                                      & \multicolumn{1}{c|}{\textbf{0.4}} & 1.32 \\ \hline
\end{tabular}
\end{table}

\begin{table}[t!]
\centering
\caption{SPL - Success weighted by Path Length}
\label{tab:length}
\begin{tabular}{|ccccc|}
\hline
Environment                & \begin{tabular}[c]{@{}c@{}}Room\\ (no object)\end{tabular} & \begin{tabular}[c]{@{}c@{}}Room\\ (with object)\end{tabular} & Office                             & Tunnel        \\ \hline
\multicolumn{1}{|c|}{A*}   & \multicolumn{1}{c|}{\textbf{0.97}}                          & \multicolumn{1}{c|}{-}                                       & \multicolumn{1}{c|}{0.41}          & 0.81          \\ \hline
\multicolumn{1}{|c|}{RRT*} & \multicolumn{1}{c|}{0.84}                                  & \multicolumn{1}{c|}{-}                                       & \multicolumn{1}{c|}{0.37}          & 0.77          \\ \hline
\multicolumn{1}{|c|}{BIT*} & \multicolumn{1}{c|}{0.83}                                  & \multicolumn{1}{c|}{-}                                       & \multicolumn{1}{c|}{0.39}          & 0.80          \\ \hline
\multicolumn{1}{|c|}{R-NAMO} & \multicolumn{1}{c|}{0.96}                                  & \multicolumn{1}{c|}{0.58}                                    & \multicolumn{1}{c|}{-}             & -             \\ \hline
\multicolumn{1}{|c|}{LAMB} & \multicolumn{1}{c|}{0.96}                                  & \multicolumn{1}{c|}{0.61}                                    & \multicolumn{1}{c|}{-}             & -             \\ \hline
\multicolumn{1}{|c|}{FAR}  & \multicolumn{1}{c|}{0.96}                                  & \multicolumn{1}{c|}{-}                                       & \multicolumn{1}{c|}{0.43}          & 0.79          \\ \hline
\multicolumn{1}{|c|}{Ours} & \multicolumn{1}{c|}{0.96}                                  & \multicolumn{1}{c|}{\textbf{0.70}}                           & \multicolumn{1}{c|}{\textbf{0.86}} & \textbf{0.97} \\ \hline
\end{tabular}
\vspace{-1.2cm}
\end{table}

\subsection{System-level Comparison}
To illustrate the potentiality and reliability of our algorithm in field application scenarios, we compare the systematic navigation performance with FAR. The reasons for comparing these two algorithms are: 1) the superiority of FAR over other path search methods has been demonstrated in \cite{yang2022far}, and 2) the offline implementation of R-NAMO and LAMB make them fail to be applicable navigation systems for comparison.
We choose Office and Tunnel for test because they are more close to the field applications. The evaluation metrics included travel distance, navigation time, and success rate for various sub-tasks, with both systems configured identically in terms of local map size, resolution, and update frequency.

\begin{figure*}[!t]
    \centering
    \begin{minipage}{0.48\textwidth}
        \centering
        \includegraphics[width=0.9\textwidth]{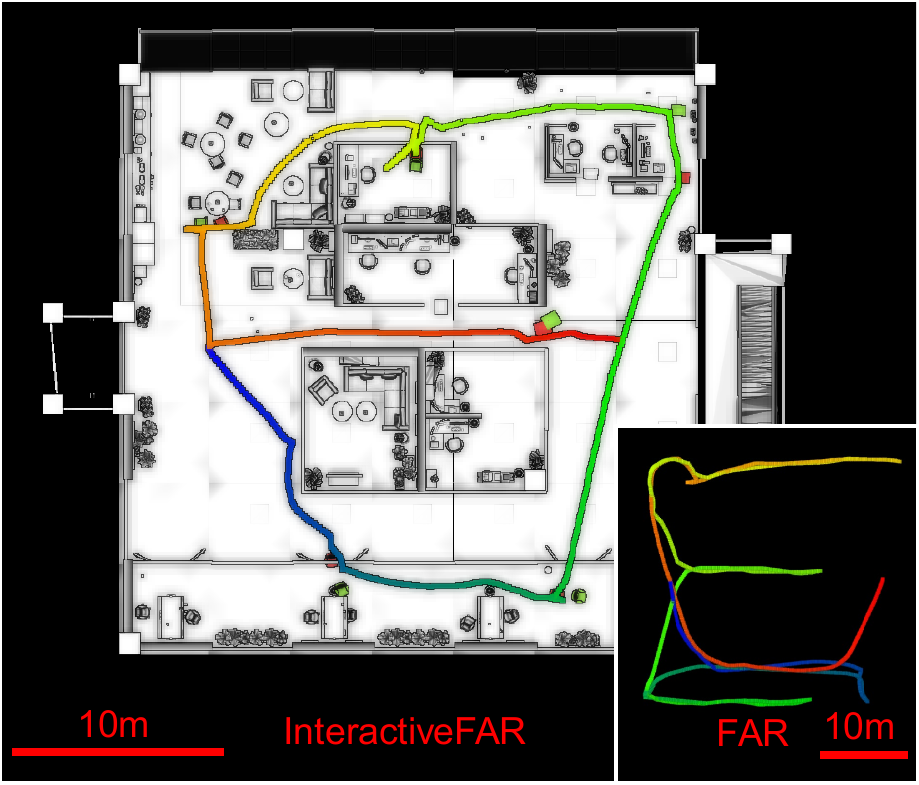} 
        \captionsetup{font={small}}
        \caption{The resulting map and trajectories of system-level experiment in Office.}
        \label{fig:office}
    \end{minipage}\hfill
    \begin{minipage}{0.48\textwidth}
        \centering
        \includegraphics[width=0.9\textwidth]{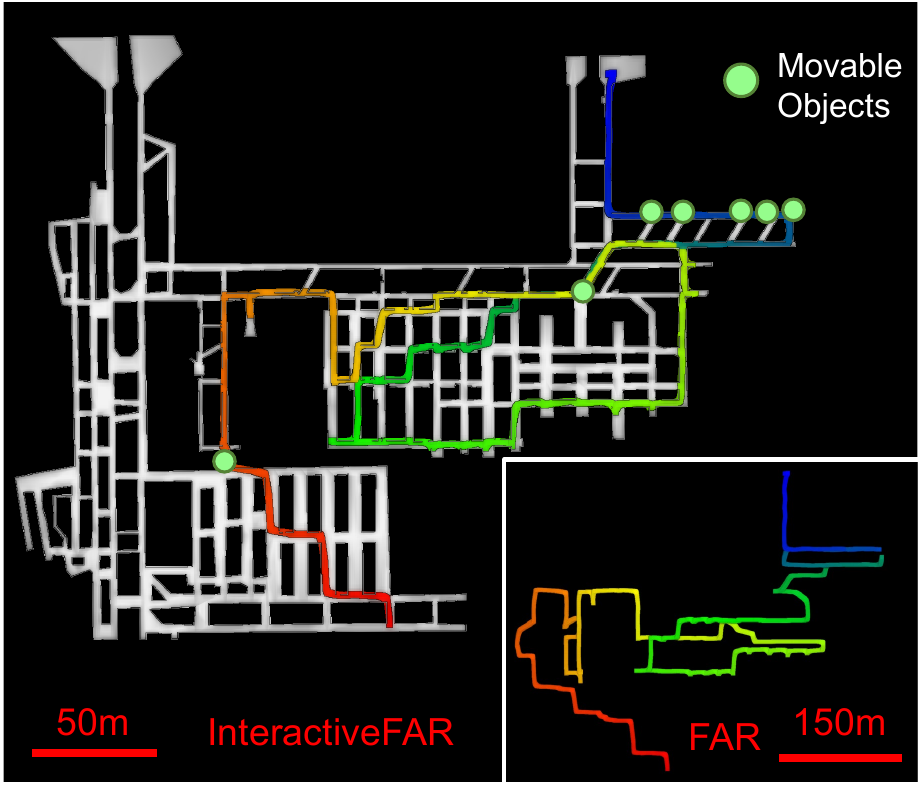} 
        \captionsetup{font={small}}
        \caption{The resulting map and trajectories of system-level experiment in Tunnel.}
        \label{fig:tunnel}
    \end{minipage}
\vspace{-0.7cm}
\end{figure*}

The results are shown in Fig. \ref{fig:office}-\ref{fig:tunnel} and Table. \ref{tab:system}. Both algorithm achieves satisfying performance in terms of success rate. However, due to the lack of interactive navigation ability, FAR planner makes more attempts to finish each sub-task, and sometimes even fail to achieve the goal, e.g. waypoint D in Fig. \ref{fig:env}(b). The results in Table. \ref{tab:system} proves that our algorithm has higher efficiency in different environments.




\begin{table*}[!t]
\caption{System-level Comparison}
\label{tab:system}
\centering
\resizebox{0.95\textwidth}{!}{%
\begin{tabular}{|c|ccc|ccc|}
\hline
Environment    & \multicolumn{3}{c|}{Office}                                                                        & \multicolumn{3}{c|}{Tunnel}                                                           \\ \hline
Metrics        & \multicolumn{1}{c|}{Travel Distance (m)}          & \multicolumn{1}{c|}{Time (s)}                & Success Rate & \multicolumn{1}{c|}{Travel Distance (m)}    & \multicolumn{1}{c|}{Time (s)}         & Success Rate \\ \hline
FAR            & \multicolumn{1}{c|}{232.56}          & \multicolumn{1}{c|}{360.34}          & 4/5          & \multicolumn{1}{c|}{1489.63}    & \multicolumn{1}{c|}{1911.45}   & 3/3          \\ \hline
InteractiveFAR & \multicolumn{1}{c|}{\textbf{109.72}} & \multicolumn{1}{c|}{\textbf{242.89}} & \textbf{5/5} & \multicolumn{1}{c|}{\textbf{1120.91}} & \multicolumn{1}{c|}{\textbf{1486.78}} & 3/3       \\ \hline
\end{tabular}%
}
\vspace{-0.2cm}
\end{table*}

\section{Conclusion}
In this paper, we present InteractiveFAR, an interactive navigation system for complex unknown environments cluttered with movable obstacles. 
Utilizing dynamic DV-graph that integrates interactions during mapping, the system outperforms benchmarks in path searching.
The proposed interactive motion planning and adaptive replan framework helps our system manipulate movable obstacles and adjust strategies in real-time based on new sensor data.
Comprehensive experiments and benchmark comparisons validate the efficiency and potential of our system in field applications.


\bibliographystyle{IEEEtran}

\bibliography{references}

\begin{thebibliography}{10}
\providecommand{\url}[1]{#1}
\csname url@samestyle\endcsname
\providecommand{\newblock}{\relax}
\providecommand{\bibinfo}[2]{#2}
\providecommand{\BIBentrySTDinterwordspacing}{\spaceskip=0pt\relax}
\providecommand{\BIBentryALTinterwordstretchfactor}{4}
\providecommand{\BIBentryALTinterwordspacing}{\spaceskip=\fontdimen2\font plus
\BIBentryALTinterwordstretchfactor\fontdimen3\font minus \fontdimen4\font\relax}
\providecommand{\BIBforeignlanguage}[2]{{%
\expandafter\ifx\csname l@#1\endcsname\relax
\typeout{** WARNING: IEEEtran.bst: No hyphenation pattern has been}%
\typeout{** loaded for the language `#1'. Using the pattern for}%
\typeout{** the default language instead.}%
\else
\language=\csname l@#1\endcsname
\fi
#2}}
\providecommand{\BIBdecl}{\relax}
\BIBdecl

\bibitem{cao2021tare}
C.~Cao, H.~Zhu, H.~Choset, and J.~Zhang, ``Tare: A hierarchical framework for efficiently exploring complex 3d environments.'' in \emph{Robotics: Science and Systems}, vol.~5, 2021.

\bibitem{shahidzadeh2023actexplore}
A.-H. Shahidzadeh, S.~J. Yoo, P.~Mantripragada, C.~D. Singh, C.~Ferm{\"u}ller, and Y.~Aloimonos, ``Actexplore: Active tactile exploration on unknown objects,'' \emph{arXiv preprint arXiv:2310.08745}, 2023.

\bibitem{farahnakian2021towards}
F.~Farahnakian, L.~Koivunen, T.~M{\"a}kil{\"a}, and J.~Heikkonen, ``Towards autonomous industrial warehouse inspection,'' in \emph{2021 26th International Conference on Automation and Computing (ICAC)}.\hskip 1em plus 0.5em minus 0.4em\relax IEEE, 2021, pp. 1--6.

\bibitem{zhou2022swarm}
X.~Zhou, X.~Wen, Z.~Wang, Y.~Gao, H.~Li, Q.~Wang, T.~Yang, H.~Lu, Y.~Cao, C.~Xu \emph{et~al.}, ``Swarm of micro flying robots in the wild,'' \emph{Science Robotics}, vol.~7, no.~66, p. eabm5954, 2022.

\bibitem{wilfong1988motion}
G.~Wilfong, ``Motion planning in the presence of movable obstacles,'' in \emph{Proceedings of the fourth annual symposium on Computational geometry}, 1988, pp. 279--288.

\bibitem{stilman2008planning}
M.~Stilman and J.~Kuffner, ``Planning among movable obstacles with artificial constraints,'' \emph{The International Journal of Robotics Research}, vol.~27, no. 11-12, pp. 1295--1307, 2008.

\bibitem{levihn2014locally}
M.~Levihn, M.~Stilman, and H.~Christensen, ``Locally optimal navigation among movable obstacles in unknown environments,'' in \emph{2014 IEEE-RAS International Conference on Humanoid Robots}.\hskip 1em plus 0.5em minus 0.4em\relax IEEE, 2014, pp. 86--91.

\bibitem{muguira2023visibility}
J.~Muguira-Iturralde, A.~Curtis, Y.~Du, L.~P. Kaelbling, and T.~Lozano-P{\'e}rez, ``Visibility-aware navigation among movable obstacles,'' in \emph{2023 IEEE International Conference on Robotics and Automation (ICRA)}.\hskip 1em plus 0.5em minus 0.4em\relax IEEE, 2023, pp. 10\,083--10\,089.

\bibitem{xia2020interactive}
F.~Xia, W.~B. Shen, C.~Li, P.~Kasimbeg, M.~E. Tchapmi, A.~Toshev, R.~Mart{\'\i}n-Mart{\'\i}n, and S.~Savarese, ``Interactive gibson benchmark: A benchmark for interactive navigation in cluttered environments,'' \emph{IEEE Robotics and Automation Letters}, vol.~5, no.~2, pp. 713--720, 2020.

\bibitem{zeng2021pushing}
K.-H. Zeng, L.~Weihs, A.~Farhadi, and R.~Mottaghi, ``Pushing it out of the way: Interactive visual navigation,'' in \emph{Proceedings of the IEEE/CVF Conference on Computer Vision and Pattern Recognition}, 2021, pp. 9868--9877.

\bibitem{suzuki1985topological}
S.~Suzuki \emph{et~al.}, ``Topological structural analysis of digitized binary images by border following,'' \emph{Computer vision, graphics, and image processing}, vol.~30, no.~1, pp. 32--46, 1985.

\bibitem{douglas1973algorithms}
D.~H. Douglas and T.~K. Peucker, ``Algorithms for the reduction of the number of points required to represent a digitized line or its caricature,'' \emph{Cartographica: the international journal for geographic information and geovisualization}, vol.~10, no.~2, pp. 112--122, 1973.

\bibitem{yang2022far}
F.~Yang, C.~Cao, H.~Zhu, J.~Oh, and J.~Zhang, ``Far planner: Fast, attemptable route planner using dynamic visibility update,'' in \emph{2022 IEEE/RSJ International Conference on Intelligent Robots and Systems (IROS)}.\hskip 1em plus 0.5em minus 0.4em\relax IEEE, 2022, pp. 9--16.

\bibitem{rodriguez2012path}
A.~Rodriguez and M.~T. Mason, ``Path connectivity of the free space,'' \emph{IEEE Transactions on Robotics}, vol.~28, no.~5, pp. 1177--1180, 2012.

\bibitem{lynch1996stable}
K.~M. Lynch and M.~T. Mason, ``Stable pushing: Mechanics, controllability, and planning,'' \emph{The international journal of robotics research}, vol.~15, no.~6, pp. 533--556, 1996.

\bibitem{mason1986mechanics}
M.~T. Mason, ``Mechanics and planning of manipulator pushing operations,'' \emph{The International Journal of Robotics Research}, vol.~5, no.~3, pp. 53--71, 1986.

\bibitem{hogan2018reactive}
F.~R. Hogan, E.~R. Grau, and A.~Rodriguez, ``Reactive planar manipulation with convex hybrid mpc,'' in \emph{2018 IEEE International Conference on Robotics and Automation (ICRA)}.\hskip 1em plus 0.5em minus 0.4em\relax IEEE, 2018, pp. 247--253.

\bibitem{doshi2020hybrid}
N.~Doshi, F.~R. Hogan, and A.~Rodriguez, ``Hybrid differential dynamic programming for planar manipulation primitives,'' in \emph{2020 IEEE International Conference on Robotics and Automation (ICRA)}.\hskip 1em plus 0.5em minus 0.4em\relax IEEE, 2020, pp. 6759--6765.

\bibitem{houle1985computing}
M.~E. Houle and G.~T. Toussaint, ``Computing the width of a set,'' in \emph{Proceedings of the first annual symposium on Computational geometry}, 1985, pp. 1--7.

\bibitem{dolgov2010path}
D.~Dolgov, S.~Thrun, M.~Montemerlo, and J.~Diebel, ``Path planning for autonomous vehicles in unknown semi-structured environments,'' \emph{The international journal of robotics research}, vol.~29, no.~5, pp. 485--501, 2010.

\bibitem{cao2022icra}
C.~Cao, H.~Zhu, F.~Yang, Y.~Xia, H.~Choset, J.~Oh, and J.~Zhang, ``Autonomous exploration development environment and the planning algorithms,'' in \emph{2022 International Conference on Robotics and Automation (ICRA)}, 2022, pp. 8921--8928.

\bibitem{zehra2020comparative}
F.~Zehra, M.~Javed, D.~Khan, and M.~Pasha, ``Comparative analysis of c++ and python in terms of memory and time. 2020,'' \emph{Preprints.[Google Scholar]}, 2020.

\bibitem{anderson2018evaluation}
P.~Anderson, A.~Chang, D.~S. Chaplot, A.~Dosovitskiy, S.~Gupta, V.~Koltun, J.~Kosecka, J.~Malik, R.~Mottaghi, M.~Savva \emph{et~al.}, ``On evaluation of embodied navigation agents,'' \emph{arXiv preprint arXiv:1807.06757}, 2018.

\end{thebibliography}


\end{document}